%% file: nips2021.tex
\documentclass{article}

\PassOptionsToPackage{numbers, compress}{natbib}

\usepackage[final]{neurips_2021}

\usepackage[utf8]{inputenc} 
\usepackage[T1]{fontenc}    
\usepackage{hyperref}       
\usepackage{url}            
\usepackage{booktabs}       
\usepackage{amsfonts}       
\usepackage{amsmath, amssymb}
\usepackage{nicefrac}       
\usepackage{microtype}      
\usepackage{xcolor}         
\usepackage[pdftex]{graphicx}

\title{Expert Human-Level Driving in Gran Turismo Sport \\
       Using Deep Reinforcement Learning \\
       with Image-based Representation}

\author{%
  Ryuji Imamura\thanks{Work done during internship at Sony AI Inc.}\\
  Tokyo \\
  \texttt{ry.immr@gmail.com} \\
  \And
  Takuma Seno\\
  Sony AI Inc.\\
  Tokyo \\
  \texttt{takuma.seno@sony.com} \\
  \And
  Kenta Kawamoto\\
  Sony AI Inc.\\
  Tokyo \\
  \texttt{kenta.kawamoto@sony.com} \\
  \And
  Michael Spranger\\
  Sony AI Inc.\\
  Tokyo \\
  \texttt{michael.spranger@sony.com} \\
}

\begin{document}

\maketitle

\begin{abstract}
\input{abstract.tex}
\end{abstract}

\section{Introduction}
\label{introduction}
\input{introduction.tex}

\section{Related Work}
\label{related_work}
\input{related_work.tex}

\section{Methodology}
\label{methodology}
\input{methodology.tex}

\section{Experiments}
\label{experiments}
\input{experiments.tex}

\section{Conclusion}
\label{conclusion}
\input{conclusion.tex}

\begin{ack}
We would like to thank Polyphony Digital Inc. for enabling this research. We also would like to thank Florian Fuchs from Sony AI Zurich for his prior implementation and Pete Wurman from Sony AI America for his feedback on the manuscript.
\end{ack}

{
\small

\bibliographystyle{unsrt}
\bibliography{nips2021}
}

\end{document}

%% file: abstract.tex
When humans play virtual racing games, they use visual environmental information on the game screen to understand the rules within the environments. In contrast, a state-of-the-art realistic racing game AI agent that outperforms human players does not use image-based environmental information but the compact and precise measurements provided by the environment. In this paper, a vision-based control algorithm is proposed and compared with human player performances under the same conditions in realistic racing scenarios using Gran Turismo Sport (GTS), which is known as a high-fidelity realistic racing simulator. In the proposed method, the environmental information that constitutes part of the observations in conventional state-of-the-art methods is replaced with feature representations extracted from game screen images. We demonstrate that the proposed method performs expert human-level vehicle control under high-speed driving scenarios even with game screen images as high-dimensional inputs. Additionally, it outperforms the built-in AI in GTS in a time trial task, and its score places it among the top 10\% approximately 28,000 human players.

%% file: introduction.tex
Recently, super-human performance was demonstrated in a time trial setting in Gran Turismo Sport (GTS) \cite{Fuchs_2021}, a realistic racing simulator based on real-world dynamics. The result was achieved using a neural network-based controller that has learned policies for vehicle control through reinforcement learning. This state-of-the-art method formulates the minimum-time race problem by maximizing course progression and uses a multi-layer perceptron to learn the mapping from low-dimensional observations to control commands without relying on human intervention, expert data, or explicit path planning. The controller observes all the necessary information for control, such as speed, acceleration, course curvatures, and distance to the edge of the racetrack, directly from the simulator.

In contrast, when human players play a racing game, they decide on their next action based on the environmental information obtained from the game screen, heads-up display (HUD) information on the screen, and the feedback vibration to their hand. This implies that the game screen provides enough information to control the vehicle and suggests that it is not necessary to use precise observations provided by the environment (hereinafter referred to as dedicated observation) as in conventional methods to achieve human-level performance.

A lot of research has been conducted on vehicle control using images of a road in the front of a car as observations \cite{lee2019deep, jaritz2018endtoend, li2018reinforcement}. In the context of racing games, end-to-end vehicle control has been achieved by direct perception in TORCS \cite{TORCS}, which is an open-source racing simulator. However, to the best of our knowledge, there are few reports on the performance of vehicle control under high-speed driving conditions in a high-fidelity realistic driving simulator. Jaritz et al. \cite{jaritz2018endtoend} have worked on learning policies for end-to-end vehicle control using photorealistic game screen images and the car's velocity as observations; however, the reward design was based on the distance from the center of the track and did not consider shortest path planning. Therefore, it is still unclear whether expert human-level performance can be achieved using high-dimensional environmental observations such as high-resolution photorealistic game screen images in a time trial task that requires learning and control to always take the shortest path.

This study aims to train the controller for the time trial task in GTS using game screen images and standard sensor information, such as the car's velocity and acceleration, as observations. The approach is to first learn the feature representations from the game screen images for control, and then learn the vehicle control using the representations, instead of end-to-end learning with images as observations.

In this paper, we experimented with the proposed controller to measure lap time in the time trial task in GTS and compare it with the scores of the built-in AI and human players. The controller not only outperformed the built-in AI but also performed within the top 10\% of the approximately 28,000 human players. Furthermore, we confirmed that the controller learned to drive on trajectories that just barely avoided contact with the inner wall in most corners, recognizing the position of the vehicle in relation to the track from the game screen images. This is the first vision-based control approach that has been shown, using quantitative comparison, to achieve expert human-level performance for the problem of controlling a high-speed vehicle in realistic racing scenarios.

%% file: related_work.tex
\subsection*{Autonomous Driving in Realistic Environments}
CARLA \cite{dosovitskiy2017carla} is one of the most popular simulators for evaluating autonomous driving systems in realistic environments. It provides the necessary observations for driving operations, such as cameras, distance sensors, and location information. In addition, it is open source. Such well-equipped simulators have contributed to research on autonomous driving systems \cite{Rhinehart_2019_ICCV, Liang_2018_ECCV, tang2019mfp}. However, the dynamics of CARLA are not realistic enough for racing; therefore, it can be used for trajectory planning simulations, but not for motion control simulations.

In this study, we tackle a task that requires more precise control, which is driving in racing scenarios. For control in racing scenarios, several works have reported that reinforcement learning-based approaches can perform as well as or better than humans. Fuchs et al. \cite{Fuchs_2021} introduced course-progress proxy rewards to replace the trajectory minimization problem with the progress maximization problem and achieved superhuman performance in the GTS. Song et al. \cite{song2021autonomous} proposed the application of curriculum learning to the task of high-speed autonomous overtaking and achieved performance comparable to that of experienced humans. Cai et al. \cite{Cai_2020} formulated drift control under high-speed driving using CARLA as a trajectory-following task using data collected from experienced drivers and achieved high generalization performance even when using various vehicle types with different physical characteristics. These studies show that state-of-the-art reinforcement learning algorithms can match or outperform humans in high-speed driving control tasks that require more precise control than urban driving tasks.

The aforementioned studies used dedicated observations to observe various measurements about the car (speedometer and distance sensor values) and the shape of the racetrack. This implies that the controller needed to observe accurate values for precise control to achieve expert human-level performance. Therefore, the controller can still be considered as having some advantage over humans.

Aiming to control in the same way that humans grasp the situation from a game screen and feed it back to the operation, research has been conducted using images rather than dedicated observations for control \cite{mnih2015human, Lample_Chaplot_2017}. CARLA provides images from in-vehicle cameras and is therefore often used in simulations for image-based autonomous driving control in urban areas using reinforcement learning \cite{codevilla2018end, pmlr-v87-mueller18a}. Li et al. \cite{li2018reinforcement} proposed a control method using TORCS based on track features extracted from driver-view. Jaritz et al. \cite{jaritz2018endtoend} proposed end-to-end image-based control learning for World Rally Championship 6, a realistic racing game. Although these image-based control methods have been proposed in racing scenarios with realistic physical characteristics, they have not been quantitatively compared to control methods for human players in time trial tasks. To the best of our knowledge, none have been compared with the performance of a large number of human players under the same conditions.

\subsection*{Representation Learning in Reinforcement Learning}
Recently, several approaches have been proposed that explicitly separate the training of environmental features from the training of control policies \cite{srinivas2020curl, stooke2021decoupling}. These approaches have been demonstrated to perform as well as or better than conventional end-to-end approaches.

One non-end-to-end image-based approach is to use many pre-collected observation images to train a network offline to map the input image to a lower-dimensional feature vector than the raw image. Lee et al. \cite{lee2019deep} trained a network to estimate the heading angle, distances to preceding cars, and distance to the centerline from images and used the output values of the network for vehicle control. Li et al. \cite{li2018reinforcement} proposed a control method that outperforms existing controllers by using features learned from images as observations for reinforcement learning. Inspired by the success of these approaches, we employ representation learning on game screen images.

%% file: methodology.tex
\begin{figure}[t]
\centering
\begin{minipage}[c]{0.6\linewidth}
\centerline{\includegraphics[width=\columnwidth]{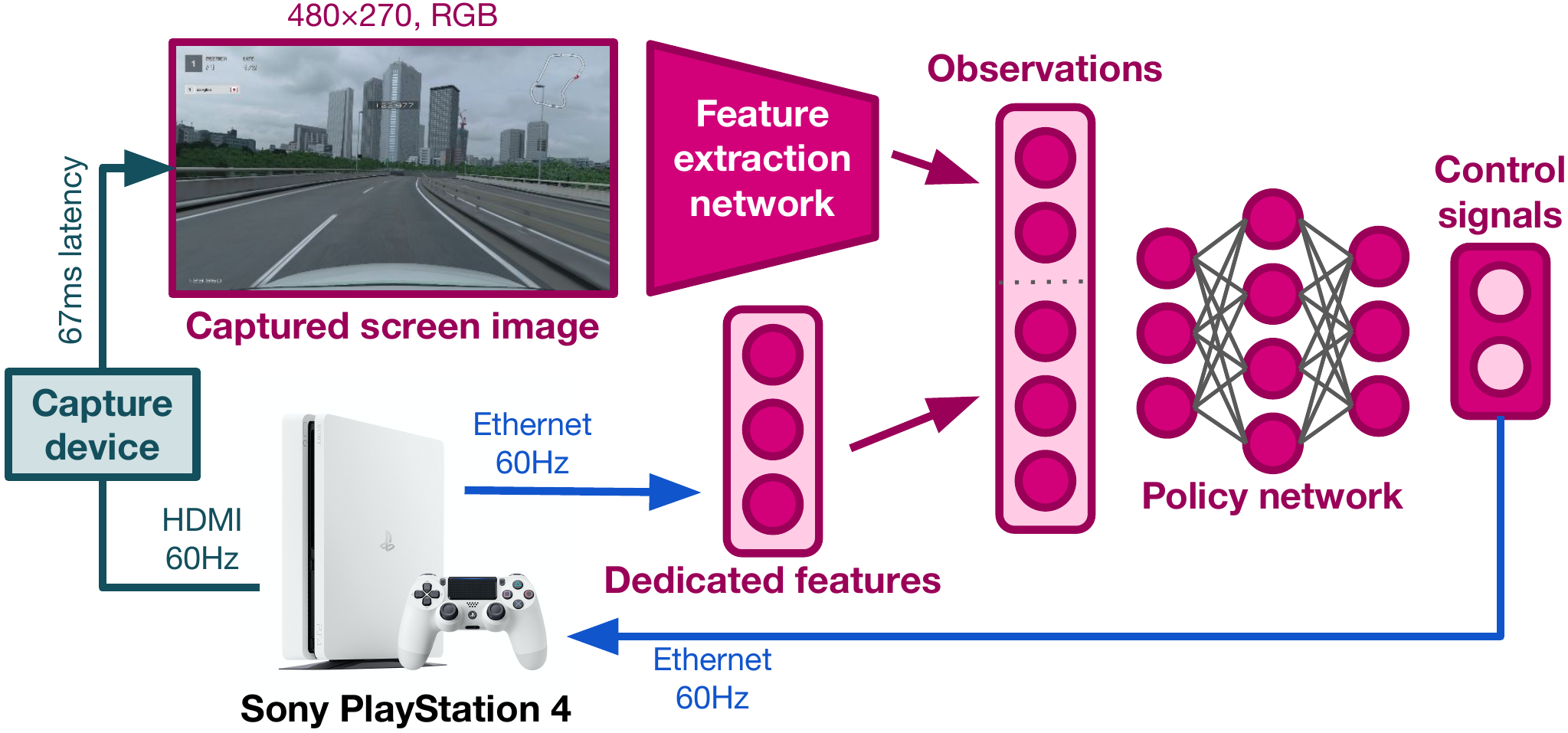}}
\end{minipage}
\begin{minipage}[c]{0.35\linewidth}
\centerline{\includegraphics[width=\columnwidth]{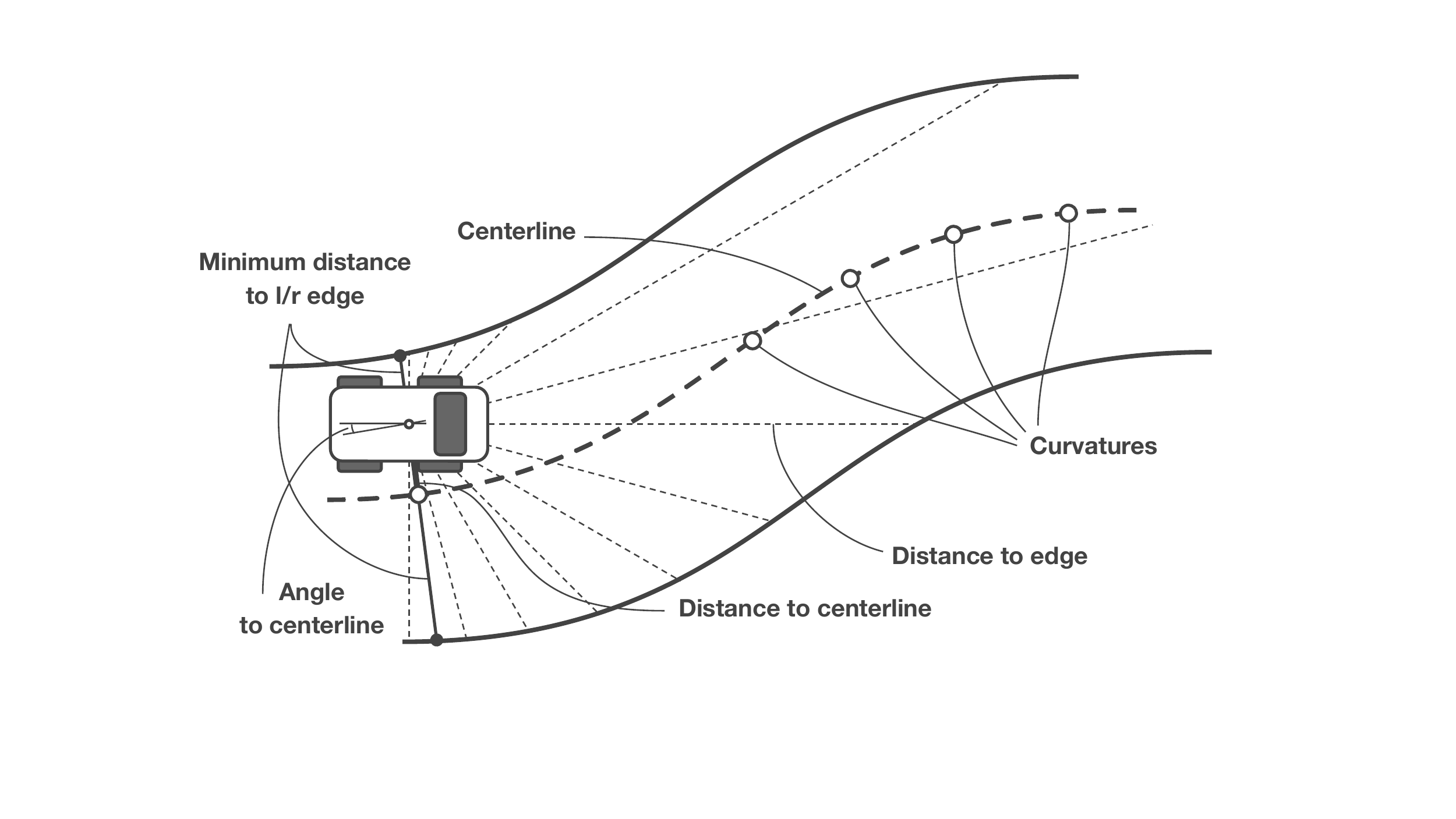}}
\end{minipage}
\caption{\textbf{(Left)} System overview of our proposed vision-based controller for the time trial in Gran Turismo Sport. Dedicated features containing standard sensor information such as a car's velocity and acceleration from the simulator on PS4 and embedded environmental representation extracted from captured game screen images are fed to the policy network to determine output signal. \textbf{(Right)} A visual illustration of objective variables in Phase 1.}
\label{fig_overview_and_environmental_features}
\end{figure}

The goal of this study is to establish precise vehicle control under high-speed driving using features extracted from game screen images as environmental observations. In previous research \cite{Fuchs_2021}, \cite{song2021autonomous}, the angle to the racetrack, the distance from the edge and center of the racetrack, and curvature values measured by in-vehicle sensors were used as environmental observations. A vision-based control algorithm is created by replacing these observations with embedded representations extracted from game screen images.

Figure \ref{fig_overview_and_environmental_features} (Left) shows an overview of our method. The screen output from PS4 at 60Hz has been captured using an HDMI capture device, and the embedded representation is obtained using it as input to the feature extraction network. Simultaneously, standard sensor information, such as a car's velocity and acceleration, is directly obtained from the PS4 as dedicated features at 60Hz. The embedded representation and the dedicated features are fed to the policy network as observations, and the output action signals are fed to the PS4 to control the vehicle.

The approach comprises a training network to acquire embedded representation from images (Phase 1) and a training policy network using standard sensor information such as a car's velocity and acceleration from the simulator and environmental representations extracted from the trained network (Phase 2). They are detailed in this section.

\subsection*{Phase 1: Image-Based Representation Learning}
The environmental observation space required for the control to outperform humans has already been narrowed down in \cite{Fuchs_2021} as follows: the distance from the center of the car body to the edge of the racetrack (in 15-degree increments over a range of 180 degrees in front), the minimum distance from the center of the car body to the left and right race track edges, the distance from the center of the car body to the centerline of the race track, the angle of the car body with respect to the centerline of the race track, the curvature of the race track at 10 points in front of the car, 3D linear velocity, 3D linear acceleration, angular velocity in the roll/pitch/yaw directions, a binary flag indicating contact with a wall, and the previous steering command.

Based on the results, a network that has explicitly learned to represent specific information is obtained by regression learning. As a preliminary study, we attempted to extract all of the above information from multiple consecutive game screen images, but we found it difficult to obtain differential information such as the speed and angular velocity of the car. In contrast, we found that environmental information, such as the distance from the car to the edge of the racetrack and the curvature of the racetrack, could be estimated well from a single image. It can be assumed that the screen image contains sufficient information necessary for estimating such environmental information, but intuitively, it is difficult to estimate very minute differential information such as the acceleration and angular velocity of the car based on pixel-by-pixel differences between multiple images. We distinguish between observations that can and cannot be estimated from the screen images and employ the following 27 pieces of information as objective variables for regression learning: the distance from the center of the car body to the edge of the racetrack (in 15-degree increments over a range of 180 degrees in front), the minimum distance from the center of the car body to the left and right race track edges, the distance from the center of the car body to the centerline of the race track, the angle of the car body with respect to the centerline of the race track, and the curvature of the race track at 10 points in front of the car. Figure \ref{fig_overview_and_environmental_features} (Right) illustrates these objective variables.

For this regression task, we trained with the following loss function to obtain a network that acquires embedded environmental representations for the images,
\begin{equation}
\label{eq_loss_function_for_regression}
l_{rep} = \|\mathbf{o}_{env} - \phi_{reg}(\phi_{rep}(\mathbf{I}))\|_{2}^{2},
\end{equation}
where $\mathbf{I} \in \mathbb{R}^{H\times W\times C}$ represents the input game screen image, with $H$, $W$, and $C$ being the height, width, and number of channels of the image, respectively; $\mathbf{o}_{env} \in \mathbb{R}^{D_{env}}$ represents the environmental observation vector corresponding to the input image; and $D_{env}$ is the number of elements in the vector. $\phi_{reg}(\phi_{rep}(\cdot))$ represents the entire network to be trained for this regression task. For convenience, $\phi_{rep}: \mathbb{R}^{H\times W\times C} \rightarrow \mathbb{R}^{D_{rep}}$ represents the layers up to the layer that outputs the $D_{rep}$-dimensional embedded environmental representations, and $\phi_{reg}: \mathbb{R}^{D_{rep}} \rightarrow \mathbb{R}^{D_{env}}$ represents the subsequent layers. 

We use the output of the $\phi_{rep}$ for the policy network rather than directly using the final output of the $\phi_{reg}$, i.e., the vector with the value of each element that is the target of regression learning, such as distance information. This is because the high-dimensional embedded representation immediately before the final output contains richer features than the final output, where each element is explicitly defined.

The dataset for regression learning is collected offline. By adding a fixed number of random values sampled from a uniform distribution to the operations of the built-in GTS AI to disrupt the operations, the game screen images from various points where the car could drive on the track are collected. Simultaneously, observations of the estimation target are collected at the same time as capturing the game screen image.

Using the collected data set, the network is trained by offline regression learning as described above. In the proposed method, the network using high-resolution images of various points on the track by offline regression learning can be trained. By separating the learning for representation acquisition from the online learning for control, the resolution of the input image can be increased, which may enable capturing of more detailed information in the image. In addition, the training to extract environmental features can be omitted during control learning, which is expected to reduce the training time of the policy.

Because the frame rate of GTS is 60 fps, the total inference time including subsequent processing must be within 16 ms as measured on the GPU. To satisfy this strict speed requirement, a lightweight architecture consisting of space2depth, which is a downsampling process based on the idea of sub-pixel convolution \cite{shi2016realtime}, and depth-wise separable convolution \cite{howard2017mobilenets} for the feature extraction network were adopted.
By reshaping the input image using space2depth, the inference is accelerated without losing the detail of high-resolution screen images. By replacing all convolutional operations with depth-wise separable convolution, the parameter size of the network is drastically reduced. The inference speed of this network averages around 8ms on the GPU, which satisfies the strict speed requirements for the subsequent training and inference tasks.

\subsection*{Phase 2: Reinforcement Learning using Learned Representation}
In addition to the embedded representation of the network that learned to estimate environmental information described in the previous section, the policy network observes 11 pieces of information needed to control the car directly from the simulator as dedicated features, including 3D linear velocity, 3D linear acceleration, angular velocity in the roll/pitch/yaw directions, a binary flag indicating contact with a wall, and the previous steering command. The policy network outputs the throttle/brake and steering values based on these inputs. Representing throttle and brake commands as a single scalar value, the output of the policy network is a 2-dimensional vector. Therefore, assuming that the command output of policy network $\phi_{p}: \mathbb{R}^{D_{rep}+D_{ded}} \rightarrow \mathbb{R}^{2}$ is a vector $\mathbf{a}_{t} \in \mathbb{R}^{2}$ consisting of $\delta_{t} \in [-\frac{\pi}{6}, \frac{\pi}{6}]$ representing the steering angle and $\omega_{t} \in [-1, 1]$ representing the scalar of throttle and brake signal, the policy network is represented as follows:
\begin{equation}
\label{eq_loss_function_for_regression}
\mathbf{a}_{t} = \phi_{p}(\mathbf{o}),
\end{equation}
where $\mathbf{o} \in \mathbb{R}^{D_{rep}+D_{ded}}$ is a combination of the embedded environmental representation $\mathbf{o}_{env}$ and $D_{ded}$-dimensional vector consisting of the values of each of the dedicated functions. In this study, we employ $D_{ded}=11$ pieces of information as dedicated functions, as mentioned above.

Our goal is to find a policy that minimizes lap time. Therefore, we adopt a reward based on the concept of maximizing course progress, represented by the following equation, following the design in \cite{Fuchs_2021},

\begin{equation}
\label{eq_reward_function}
r_{t} = r_{t}^{prog}-
\begin{cases}
c_{w}\|\mathbf{v}_{t}\|^2 & \text{if in contact with wall}\\
0 & \text{otherwise},
\end{cases}
\end{equation}
where the first term in (\ref{eq_reward_function}) is the reward for course progress evaluated at arbitrary time intervals. Course progress is calculated by projecting the trajectory of the car onto the centerline of the track. Since progress on the centerline is maximized by driving more inward at the corners of the track, we can assume that the design gives a higher reward for shorter paths. The second term corresponds to the penalty for contact with the wall and is proportional to vector $\mathbf{v}_{t} = [v_{x}, v_{y}, v_{z}]$, which represents the current linear velocity of the vehicle. Introducing this term prevents the policy from learning a more efficient time-saving strategy by hitting a wall. Here, $c_{w}$ represents the weight coefficient that controls the trade-off between these rewards and penalties.

The policy network is trained with the soft actor-critic (SAC) algorithm \cite{haarnoja2018soft}, following its success in \cite{Fuchs_2021}.

%% file: experiments.tex
\subsection*{Settings}
We evaluated our vision-based control algorithm in the time trial task in GTS. In this experiment, we used the Mazda Demio XD Turing '15 and the Tokyo expressway central outer loop as the target course. This is the same condition as that in one of the experiments in \cite{Fuchs_2021}, which allows us to compare the performance of our method with the scores of approximately 28,000 human players provided by Polyphony Digital Inc.

The following sections detail the experimental settings for representation learning and reinforcement learning.

\subsubsection*{Image-based Representation Learning}
\begin{figure}[t]
\centering
\begin{minipage}[c]{0.485\linewidth}
\hspace{-0.02\linewidth}
\centerline{\includegraphics[width=\linewidth]{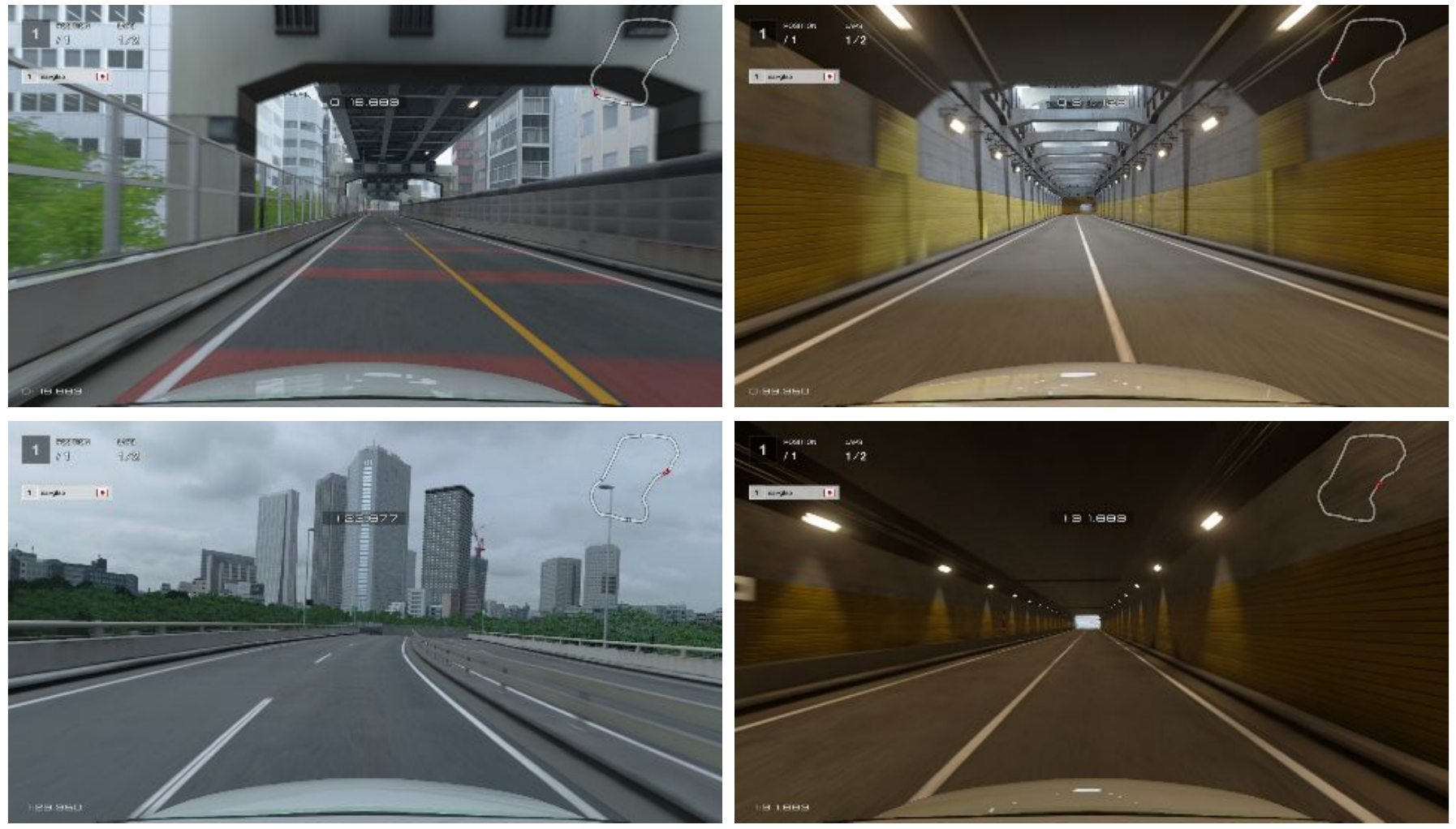}}
\end{minipage}
\begin{minipage}[c]{0.485\linewidth}
\hspace{0.005\linewidth}
\centerline{\includegraphics[width=\linewidth]{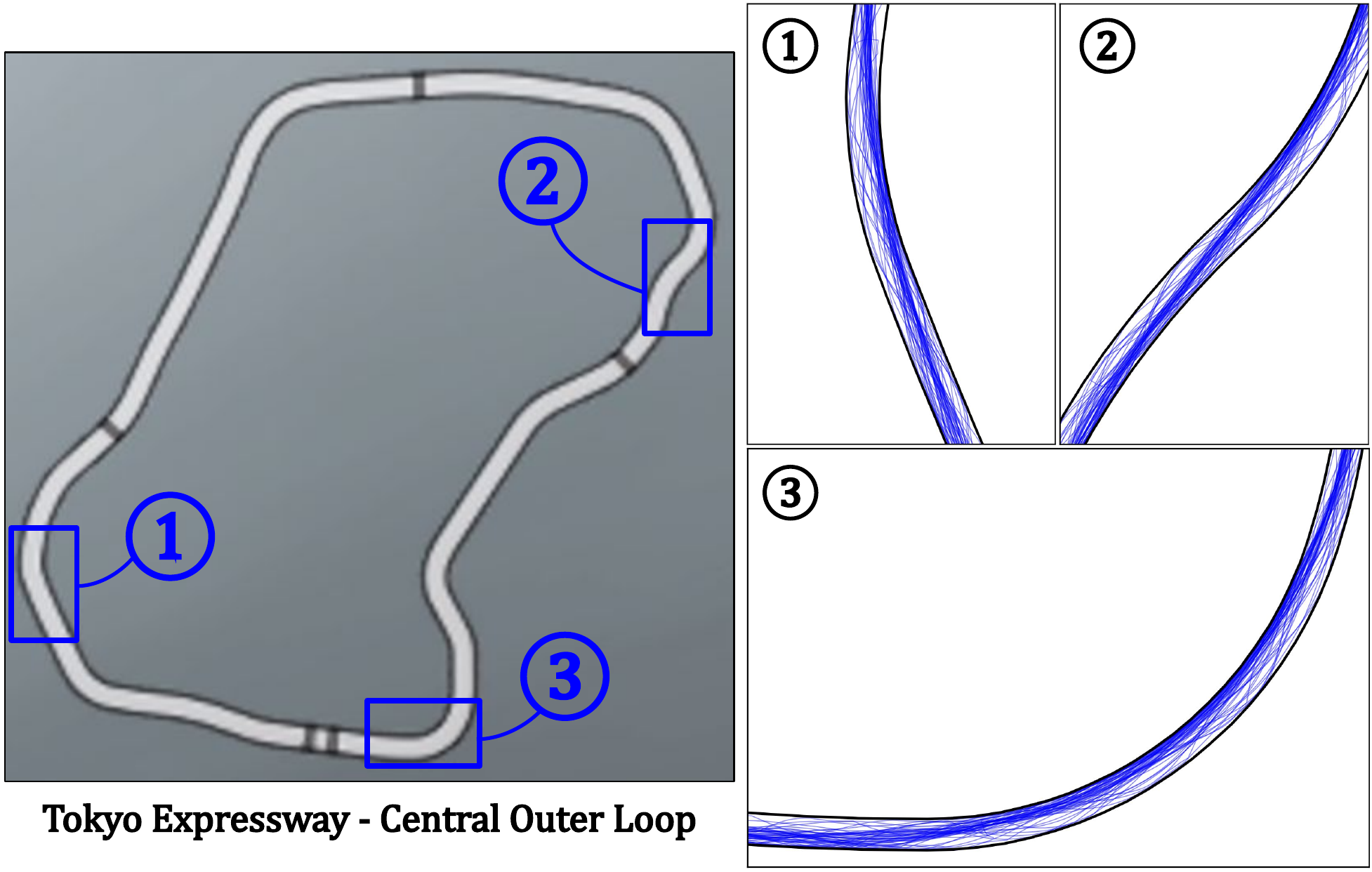}}
\end{minipage}
\caption{\textbf{(Left)} Examples of the Gran Turismo Sport game screen, driving a Mazda Demio on the "Tokyo Expressway - Central Outer Loop." \textbf{(Right)} Examples of trajectories (blue line) collected by disrupted operations to observe various points on the course.}
\label{fig_frontview_and_trajectories}
\end{figure}

For training the network, we first collected pairs of game screen images, as shown in Figure \ref{fig_frontview_and_trajectories} (Left), and observations at various points on the racetrack. Originally, HUD information, including the current speed, was displayed on the screen, but we hid it since the controller can receive it directly from the environment. We collected approximately 630,000 game screen images for 50 laps using disrupted operations described in Section 3. Figure \ref{fig_frontview_and_trajectories} (Right) shows trajectories of the car collected by the disturbed operation. The game screen images were captured at a 1280 x 720 resolution but were downsampled to a 470 x 280 resolution before being input to the network. This was adjusted based on the trade-off between inference speed and the number of features. In this experiment, $D_{env}$ was set to 27, the number of objective variables in this task, and $D_{rep}$, the vector length of the embedded environmental representation, was set at 64.

Note that only the curvature among the observations estimated by the network has a different meaning from that in previous research \cite{Fuchs_2021}. In that research, the curvature was represented by a vector consisting of values of the position of the car after 1.0 to 2.8 seconds, calculated based on the speed at that moment. In the proposed method, since training for feature extraction from images and training for car control are independent of each other, it is represented as a vector containing curvature values of the track at 10 points 40 to 120 m ahead of the current position. To keep the experimental settings as close as possible to those in the conventional method, the distance and interval for sampling the curvature values were calculated based on the approximate maximum speed of the Mazda Demio. To ensure that differences in the range of values for each observation did not affect the training priorities, each element of $\mathbf{o}_{env}$ was standardized using the mean and variance of the entire training dataset.

We prepared pairs of game images and observations obtained by operating the built-in AI without any external disruptions and used them as an evaluation dataset. We trained the network for 50 epochs and employed the weights that produced the smallest prediction error for each observation in the evaluation dataset.

\subsubsection*{Reinforcement Learning using Representation}
We trained the controller under the reward function shown in (\ref{eq_reward_function}) using the SAC algorithm. The $D_{rep}=64$-dimensional vector, which is the output of the $\phi_{rep}$ in (\ref{eq_loss_function_for_regression}), was used as the embedded environmental representation instead of the environmental information from the dedicated observations. The 75-dimensional vector, $\mathbf{o}$, which is a combination of the embedded environmental representation and the $D_{ded}=11$-dimensional vector consisting of the elements selected as in Section 3, was used as the input to the controller. Owing to the specifications of our HDMI capture device, the captured images were always 4 frames (67ms) behind the observations provided by the simulator, but we did not apply any additional processing because this delay is rather fair, considering the perceived speed of the human player. The policy network $\phi_{p}$ consisted of two 256-dimensional fully connected layers followed by a ReLU layer.

Training samples have been collected by driving 20 cars per PS4 at the same time using one PS4 \cite{song2021autonomous} or four PS4s \cite{Fuchs_2021}, but in this method, we collected samples by driving only one car using only one PS4 owing to the screen capture process. To closely match the training settings of these previous studies, we changed the starting point of the car for each trial and sampled the driving operation for 100s, updating the controller after 20 driving trials. Three seeds were used to train the controller for evaluation. For the evaluation, we let the agents run two consecutive laps and adopted the lap time of the second lap to eliminate the effect of initial speed given at the start of the evaluation according to previous research \cite{Fuchs_2021}, and the fastest time was used as the best lap score. 

\subsection*{Results}
\begin{figure}[t]
\centering
\begin{minipage}[c]{0.49\linewidth}
\centerline{\includegraphics[width=\columnwidth]{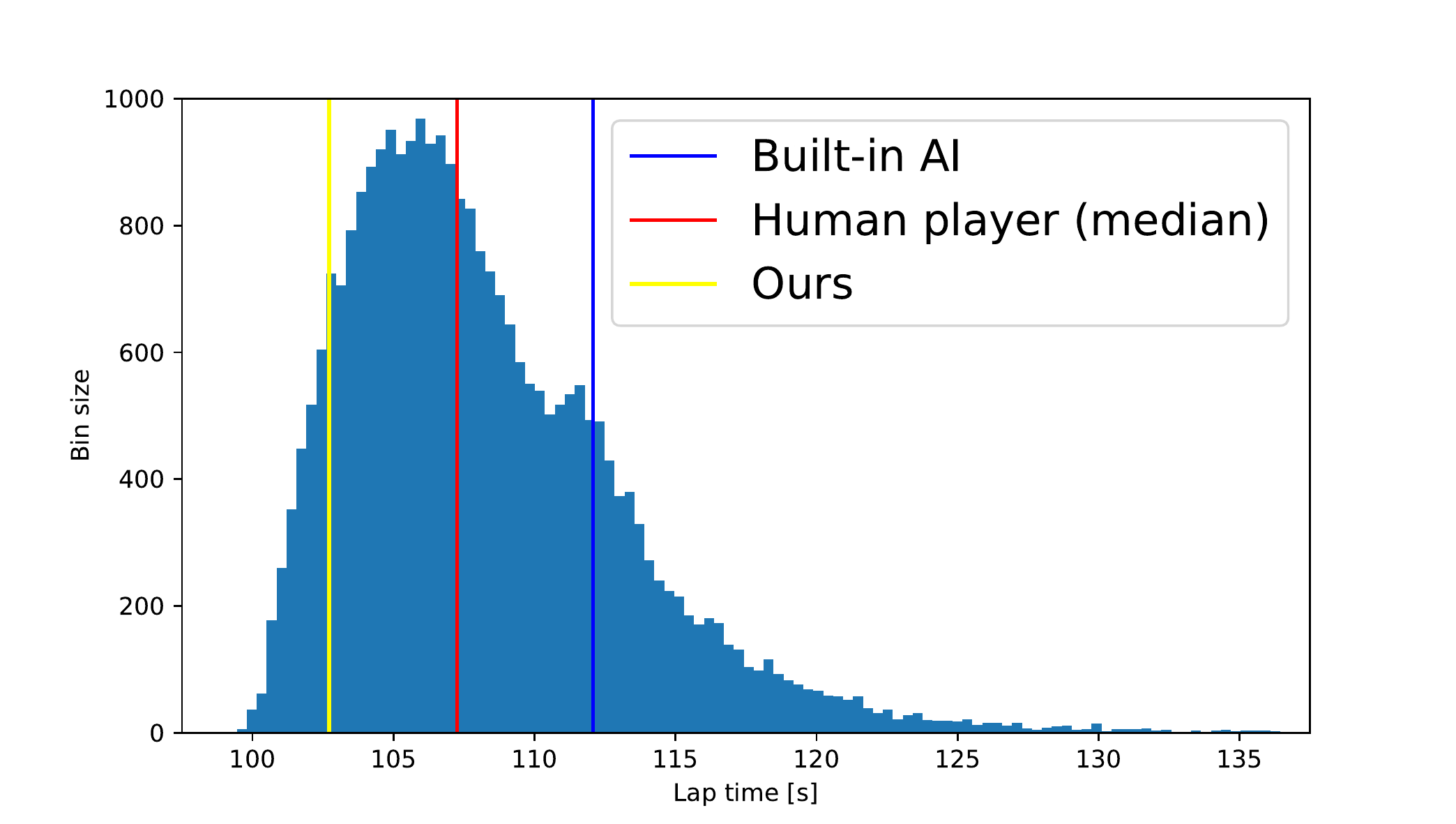}}
\end{minipage}
\begin{minipage}[c]{0.49\linewidth}
\centerline{\includegraphics[width=\columnwidth]{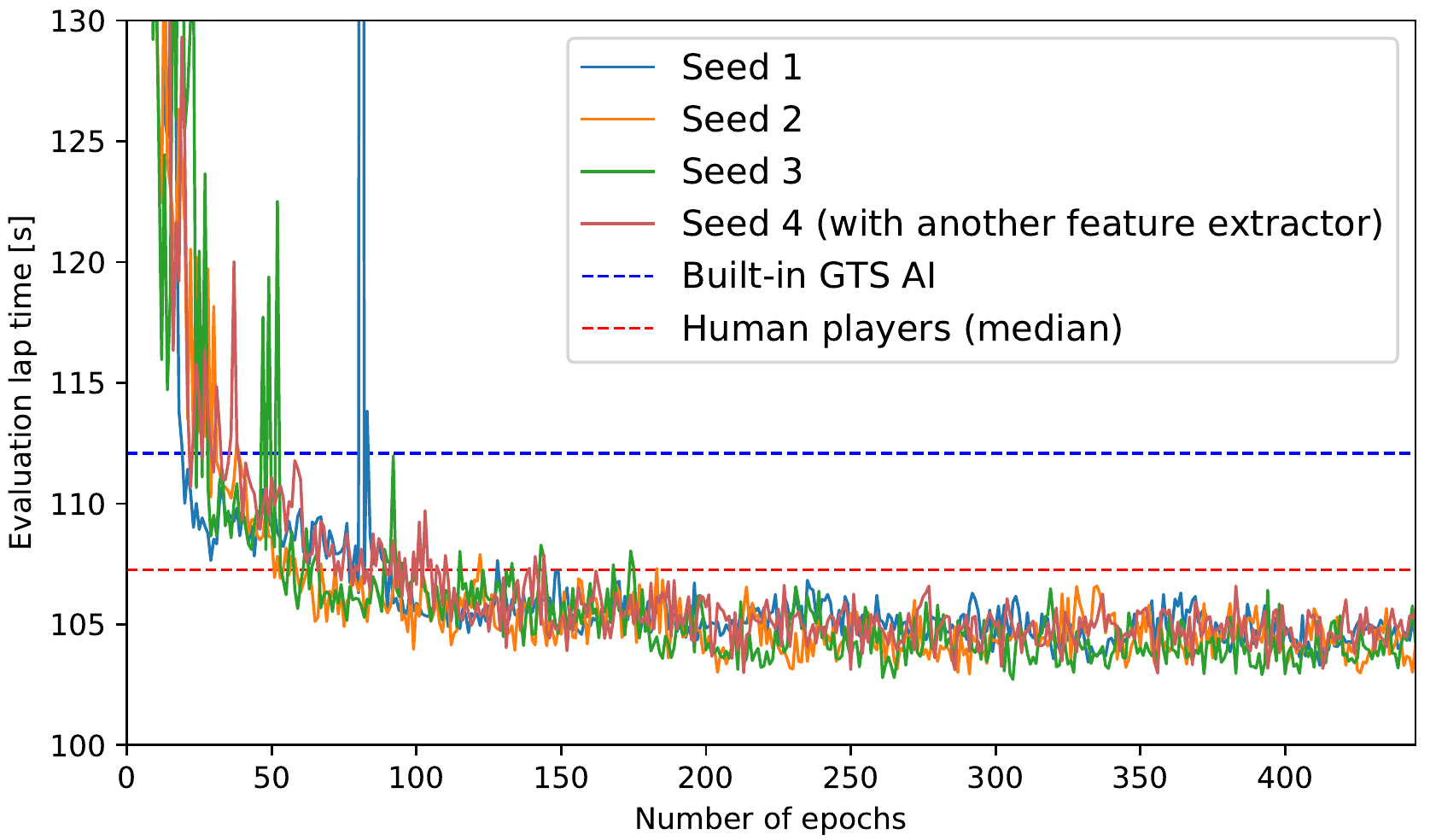}}
\end{minipage}
\caption{\textbf{(Left)} Time trial lap time comparison between over 28,000 human players (pale blue histogram), the built-in GTS AI (blue line), the median lap time of human players (red line), and our vision-based controller (yellow line). Our controller outperforms the median and mode of the scores of human players. \textbf{(Right)} Learning progress of the three agents with different initial values for the same feature extraction network and the other agent with different initial values for another feature extraction network. 1 epoch is for 20 trials of 100s each, and we update and evaluate the network parameters after the end of each epoch.}
\label{fig_histogram_and_progress}
\end{figure}

\begin{table}[t]
\caption{Time trial comparisons between the built-in GTS AI, human players, the state-of-the-art controller proposed by Fuchs et al., and our vision-based controller.}
\label{tab_results}
\centering
\begin{small}
\begin{sc}
\begin{tabular}{lcccr}
\toprule
Driver & Lap time \\
\midrule
Built-in GTS AI & 01:52.075 \\
Human players (median) & 01:47.259 \\
Human players (fastest) & 01:39.445 \\
Fuchs et al. & 01:39.408 \\
Ours & 01:42.717 \\
\bottomrule
\end{tabular}
\end{sc}
\end{small}
\end{table}

Figure \ref{fig_histogram_and_progress} (Left) and Table \ref{tab_results} present the results of the time trial in each setting. As shown in the results, the agent trained by the proposed method beat the lap time of the built-in GTS AI by 9.4s and the median score of human players by 4.5s. Furthermore, its score lies within the top 10\% of the results for scores collected from 28,000 human players. This result shows that the proposed approach can achieve expert human-level performance.

Agents trained with the conventional method, which uses dedicated observations to analyze information on the track instead of using image-based environmental information, run more optimal paths and outperform the fastest human players. Our experiments show that the performance degradation is limited to approximately 3.3 s compared to the score of the conventional method, even if dedicated observations of the environmental information are completely replaced with the representation obtained from the game screen image.

Figure \ref{fig_histogram_and_progress} (Right) shows the learning progress of three agents with different initial values for the same feature extraction network and the other agent with different initial values for a feature extraction network initialized with different values. Although there is a difference in the learning progress depending on the initial value, all agents learn the policy in approximately 400 epochs. Furthermore, it has been shown that even if the network that extracts representations from images is trained with different initial values, it performs equally well during the reinforcement learning phase. This means that the agent is robust to the representations that the network acquires.

\begin{figure}[t]
\centering
\begin{minipage}[c]{0.245\linewidth}
\centerline{\includegraphics[width=\columnwidth]{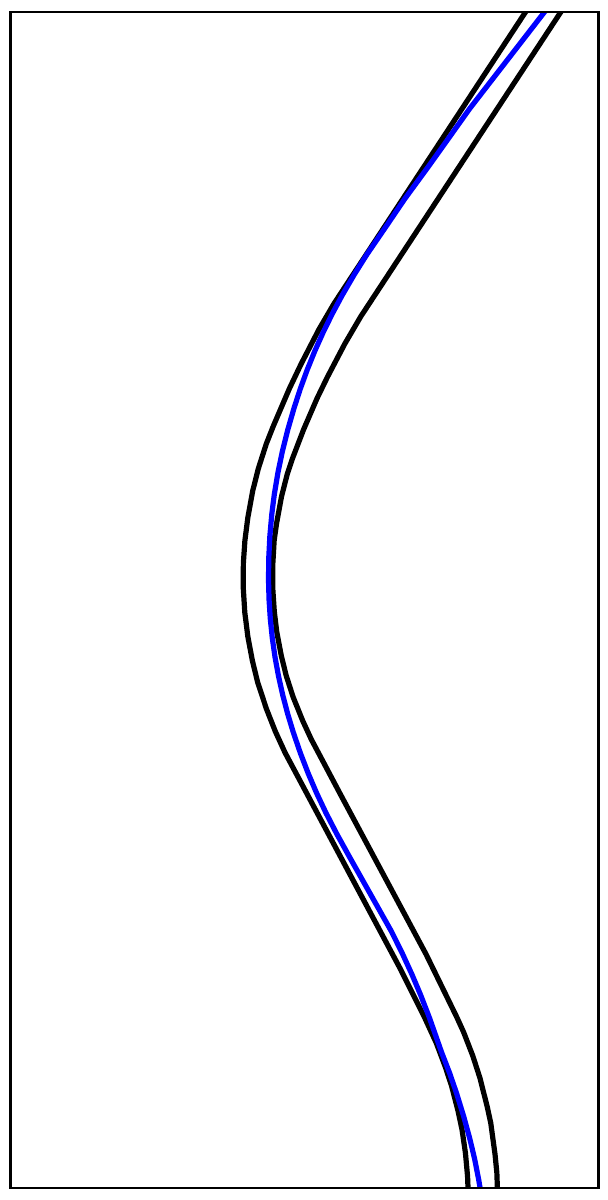}}
\end{minipage}
\begin{minipage}[c]{0.245\linewidth}
\centerline{\includegraphics[width=\columnwidth]{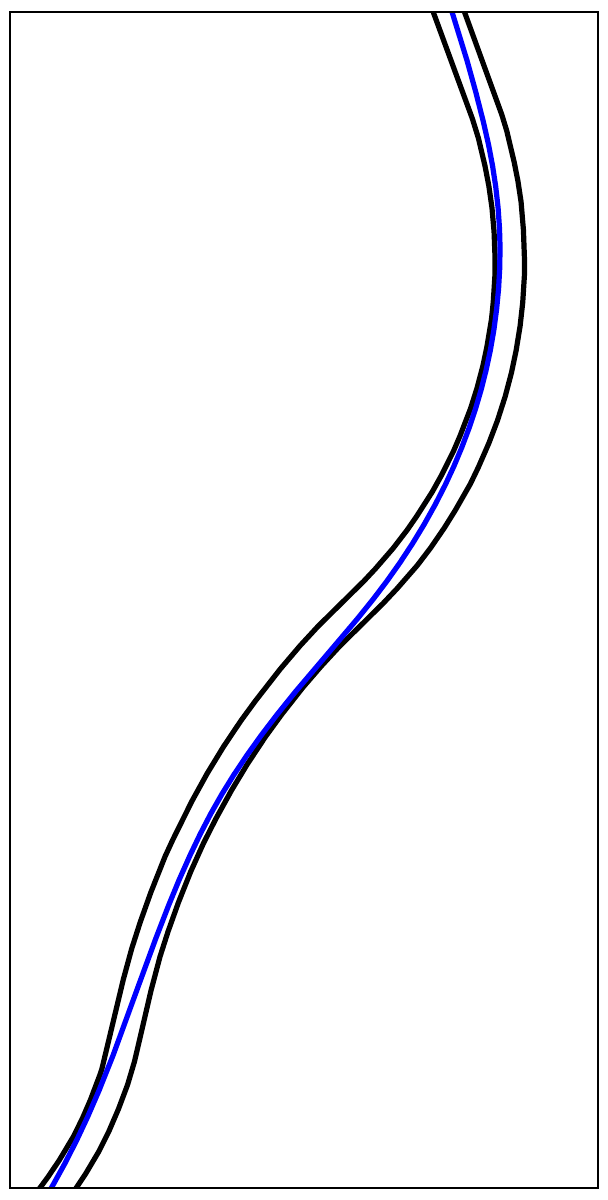}}
\end{minipage}
\begin{minipage}[c]{0.245\linewidth}
\centerline{\includegraphics[width=\columnwidth]{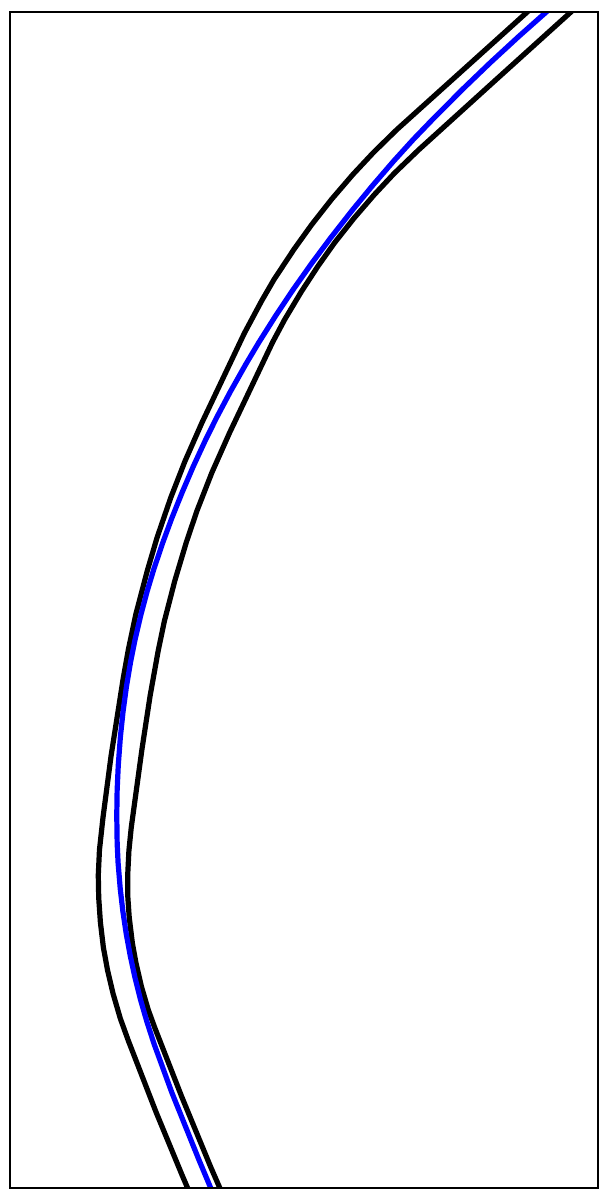}}
\end{minipage}
\begin{minipage}[c]{0.245\linewidth}
\centerline{\includegraphics[width=\columnwidth]{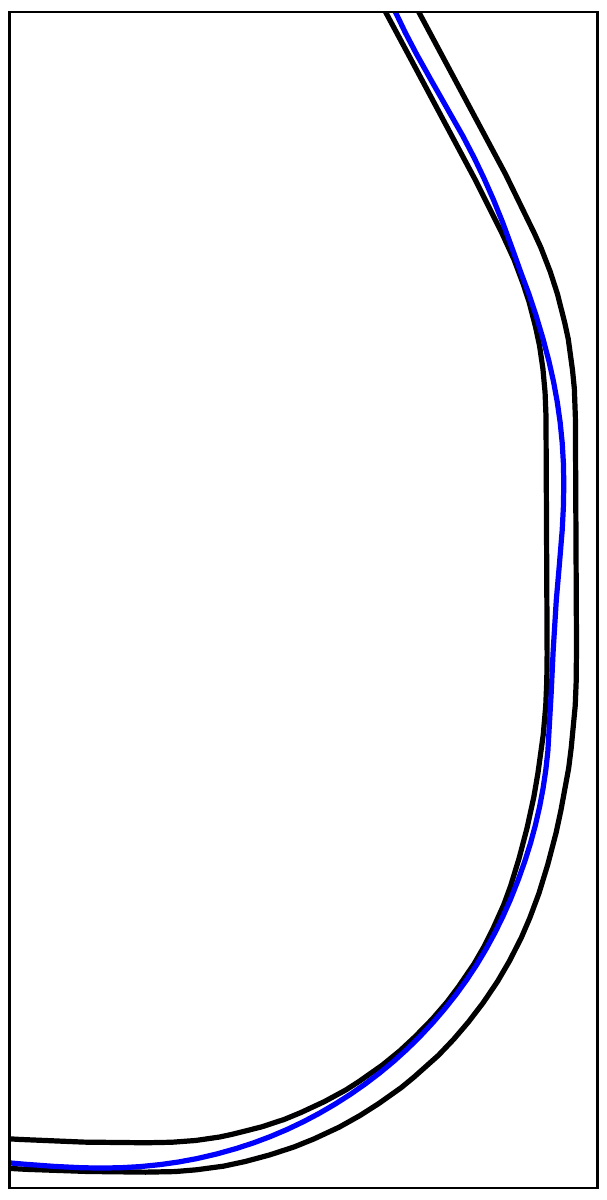}}
\end{minipage}
\caption{The trajectories of our agent during the evaluation driving in different sections (blue line). When a vehicle hits the walls, the impact causes it to not only slow down and disrupt the driving operation but also deviate from its optimal path. The trajectory of our agent is smooth and shows that our controller can drive the car without hitting a wall.}
\label{fig_eval_trajectories}
\end{figure}

Figure \ref{fig_eval_trajectories} shows the trajectories of our agent in some different types of corners in the track. Our agent accurately recognizes the position of the vehicle in relation to the track from the game screen image and drives in a near-optimal path that passes close by the walls without any collisions.

%% file: conclusion.tex
In this paper, we proposed the first vision-based control algorithm that achieves expert human-level performance on Gran Turismo Sport, a realistic racing simulator. In our approach, there are two phases: representation learning to extract the features needed to control the car from the game screen image and reinforcement learning to train the policy.

Our vision-based controller not only significantly outperformed the built-in AI but also performed within the top 10\% of 28,000 players under the same conditions in the time trial task. Moreover, compared to state-of-the-art methods that use accurate observations provided by the simulator to outperform humans, the difference was approximately 3.3s.

As a future prospect of our research, we plan to further validate the acquired feature representation. In this paper, we have not clarified that the feature extraction network does not recognize the objective variables by learning them from other features in the image. Furthermore, we intend to build a learning scheme to acquire embedded representations using multiple images to perfectly match the observation conditions with human players, since our method still observes differential information such as acceleration and angular velocity from the simulator. In addition, although we have set up an observation space for environmental information in this research based on the results of previous research \cite{Fuchs_2021}, the application and performance comparison of completely end-to-end learning as in \cite{jaritz2018endtoend} and online methods for representation learning as in \cite{srinivas2020curl, stooke2021decoupling} should be performed in a future study.